\documentclass[11pt]{article}

\usepackage[preprint]{acl}

\usepackage{times}
\usepackage{latexsym}
\usepackage{amssymb}
\usepackage{amsmath}
\usepackage{booktabs}
\usepackage[table]{xcolor}
\usepackage{multirow}
\usepackage{graphicx}
\usepackage{algorithm}
\usepackage{algorithmic}
\usepackage{dblfloatfix}
\usepackage[T1]{fontenc}

\usepackage[utf8]{inputenc}

\usepackage{microtype}

\usepackage{inconsolata}

\usepackage{graphicx}

\title{MAGIC: \underline{M}ultimodal \underline{A}lignment \& \underline{G}rounding-aware \underline{I}nstruction \underline{C}oreset for Vision-Language Models}

\author{Shristi Das Biswas \and Kaushik Roy \\
        Purdue University\\
        \tt\small sdasbisw@purdue.edu, \tt\small kaushik@purdue.edu}

\begin{document}
\maketitle
\begin{abstract}
Instruction tuning of large vision-language models (LVLMs) increasingly depends on massive multimodal corpora, yet these datasets contain samples with substantial redundancy, low visual dependency, and highly imbalanced coverage of multimodal reasoning behaviors. As a result, uniform subsampling or naive score-based selection often yields suboptimal training subsets. We introduce MAGIC, a training-free, forward-only coreset selection method designed to construct compact yet behaviorally faithful subsets for multimodal instruction tuning. MAGIC is built on three intrinsic signals extracted from a pretrained VLM: Multimodal Gain, which measures the likelihood improvement obtained from visual input; Bridging Relevance, which captures the sharpness of answer-token grounding over visual tokens; and Skill-Neuron Signatures, which characterize the functional computation elicited by each sample via top-activated feed-forward neurons. MAGIC combines these signals in a three-stage pipeline: filtering low-gain examples, ranking candidates by a normalized quality objective, and performing bucket-wise budget allocation over discrete neuron signatures to preserve latent multimodal skill coverage. This formulation avoids backpropagation, auxiliary selector training, and expensive clustering in continuous activation spaces, while remaining efficient and easily deployable in existing VLMs. Across LLaVA-665K and Vision-Flan datasets, and transfer settings to large target models, LLaVA-1.5-7B and -13B, MAGIC consistently improves over strong baselines under matched 20\% budgets: it achieves 100.3\% relative performance to full finetuning on LLaVA-665K and 101.6\% relative performance on Vision-Flan-186K, while yielding a 73.7\% reduction in wall-clock run time. Overall, our results establish that activation-aware behavioral coverage, when coupled with multimodal utility and grounding strength, provides a principled and practical foundation for VLM coreset selection.
\end{abstract}

\section{Introduction}
\begin{figure}[]
\centering
    \includegraphics[width=1.0\columnwidth]{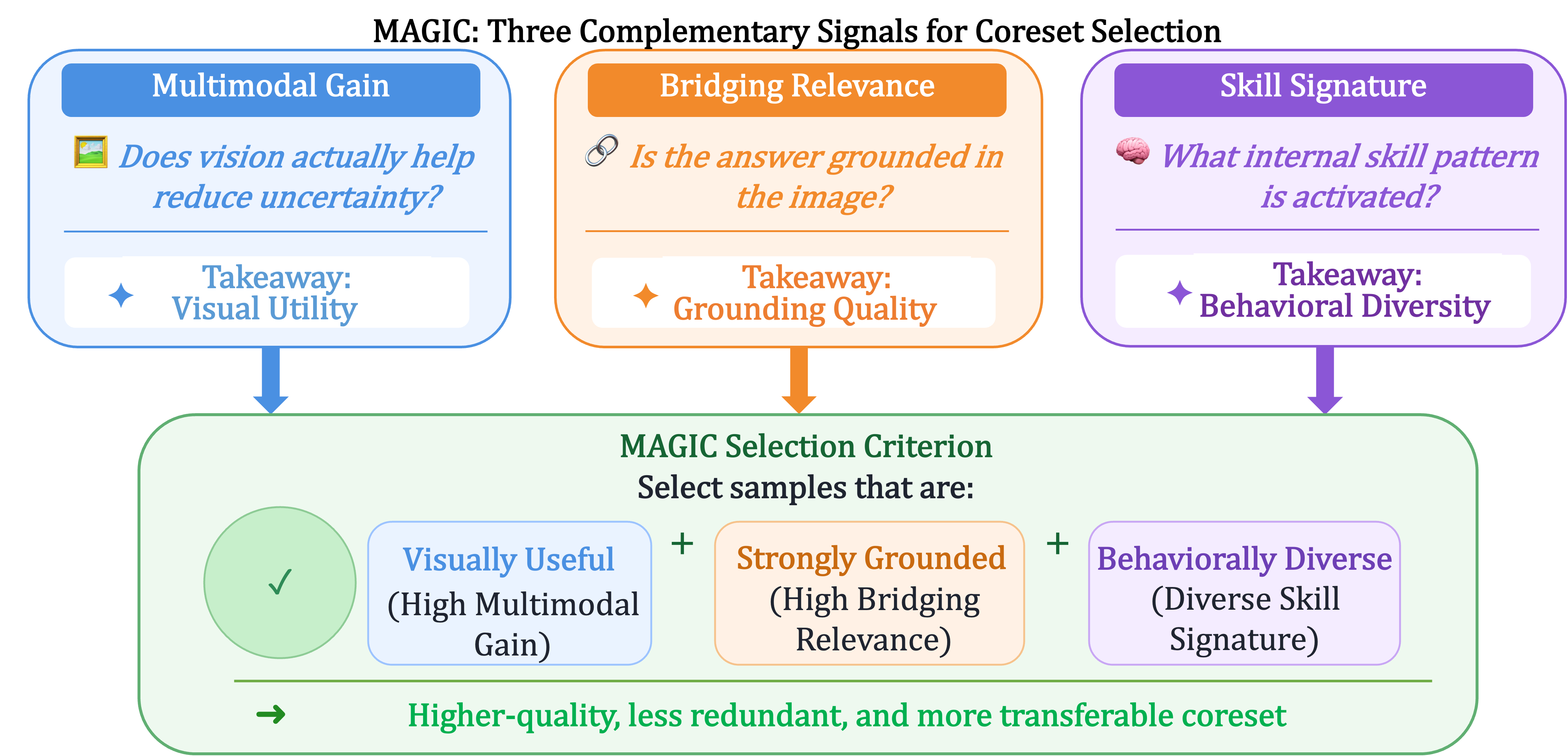} 
    \vspace{-17pt}
    \caption{MAGIC characterizes each training sample using three complementary signals: Multimodal Gain; Bridging Relevance; and Skill Signature. These signals are combined to prioritize samples that are visually useful, strongly grounded, and behaviorally diverse for visual-language coreset selection.
    }
    \vspace{-8pt}
    \label{fig:intro}
\end{figure}

Large vision-language models (LVLMs) have recently achieved strong performance across a wide range of multimodal tasks~\cite{dai2023instructblip,alayrac2022flamingo, yu2023visually, radford2021learning,zhu2023minigpt}. A standard LVLM pipeline typically consists of large-scale image-text pretraining followed by visual instruction tuning, where the model is adapted to follow task-oriented instructions. While this second stage is crucial for instruction-following ability, it has become increasingly expensive due to the rapid growth of visual instruction datasets. Moreover, these corpora are often highly redundant, contain many weakly grounded samples that require limited visual understanding, and exhibit severely imbalanced coverage of latent multimodal behaviors, making full-data tuning both costly and inefficient. Consequently, there is growing interest in multimodal data selection methods that can construct compact training subsets while preserving downstream performance.

Existing multimodal data selection methods largely inherit ideas from the broader coreset and instruction-tuning literature, including scalar score-based ranking~\cite{paul2021deep, chen2024your, du2023mods, zhou2023lima}, redundancy pruning~\cite{abbas2023semdedup}, gradient-based selection~\cite{wu2024icons, liu2025less, tan2023data}, and clustering-based diversity promotion~\cite{lee2024concept,yu2025mastering}. However, these approaches face two major limitations in the visual instruction setting. First, single-score methods often over-select narrow modes of the dataset, since highly ranked samples may repeatedly reflect the same task pattern, concept type, or response style. Second, stronger selection methods often incur substantial overhead by relying on backward passes, auxiliary selector training, or expensive clustering in continuous activation spaces. Prior work has shown that overlooking sample relationships and latent data structure can lead to poor sample uniqueness or representativeness, ultimately limiting generalization~\cite{yu2025mastering}.

In this paper, we argue that effective multimodal coreset selection should satisfy three desiderata: selected samples should exhibit strong \emph{multimodal utility}, be strongly \emph{grounded} in visual evidence, and sufficiently \emph{diverse} in the latent computations they induce. Existing methods typically address only part of this problem. Utility-based ranking captures sample quality but not coverage, while diversity-oriented methods improve spread but do not ensure that selected examples are both informative and visually grounded.

To address these limitations, we introduce MAGIC, a training-free, forward-only coreset selection method for visual instruction tuning. MAGIC is built on three intrinsic signals extracted from a pretrained VLM, as seen in Fig.~\ref{fig:intro}: \textbf{Multimodal Gain}, which measures improvement in the model’s response likelihood when multimodal inputs are present relative to a unimodal text-only forward pass; \textbf{Bridging Relevance}, which measures the concentration of answer-token attention over visual tokens; and a \textbf{Skill-Neuron Signature}, a discrete fingerprint formed from top-activated feed-forward neurons induced by a sample. Together, these signals characterize not only whether a sample is \emph{useful for multimodal learning}, but also what \emph{latent computation} it elicits inside the model. MAGIC combines these signals in a three-stage pipeline. It first filters weakly multimodal samples, then ranks the remaining candidates with a normalized quality score combining Multimodal Gain and Bridging Relevance scores, and finally allocates budget across discrete Skill-Signature buckets to preserve behavioral coverage. 

We evaluate MAGIC on two visual instruction tuning corpora, LLaVA-665K~\cite{liu2024improved} and Vision-Flan~\cite{xu2024vision}, using LLaVA-1.5-7B as the primary target model and further assessing transfer to the larger LLaVA-1.5-13B. Under matched 20\% selection budgets, MAGIC achieves 100.3\% relative performance with respect to full-data finetuning on LLaVA-665K and 101.6\% on Vision-Flan-186K, while retaining 99.3\% relative performance when transferred to LLaVA-1.5-13B. Moreover, MAGIC not only reduces end-to-end wall-clock runtime by 73.7\% relative to full-data finetuning, yielding a substantially improved compute-performance trade-off, but also improves generalization to unseen tasks. These results show that activation-aware behavioral coverage, combined with multimodal utility and grounding strength, provides a principled and efficient foundation for multimodal coreset selection. Overall, our contributions are threefold: \textbf{(i)} we propose MAGIC, a training-free, and scalable coreset selection framework for visual instruction tuning that unifies multimodal utility, visual grounding, and activation-aware behavioral diversity; \textbf{(ii)} we introduce Multimodal Gain, Bridging Relevance, and Skill-Neuron Signature bucketing as intrinsic sample-level criteria for coreset construction, without any backward passes, proxy selector training, or expensive continuous activation-space clustering; and \textbf{(iii)} we demonstrate that MAGIC delivers consistent improvements in data efficiency at just 20\% coreset selection and yields end-to-end compute-performance trade-offs across two distinct VIT datasets, LLaVA-665K and Vision-Flan, and different VLM target models, while substantially reducing total wall-clock run time by $~74\%$. We further demonstrate improved generalization to unseen tasks, despite a 5$\times$ reduction in training data.




\section{Related Works}


Coreset selection aims to form a smaller training set whose optimization behavior and downstream performance closely match those obtained from the complete data corpus. Data selection methods can be categorized based on the types of information they utilize for selection~\cite{hammoudeh2024training}. Representation-based approaches~\cite{abbas2023semdedup,lee2024concept, yu2025mastering} leverage embeddings or latent features to capture sample structure and similarity. In particular, ~\cite{lee2024concept} clusters data based on representations associated with concept-skill compositions, while~\cite{yu2025mastering} proposes a collaborative framework combining spectral analysis and clustering-derived sample relationships for multimodal data valuation. Loss-trajectory-based methods~\cite{mindermann2022prioritized} prioritize data points that contribute most significantly to reducing generalization error over training. Gradient-based techniques~\cite{xia2024less,wu2024icons,paul2021deep,deng2024influential, liu2025less} select data based on gradient information or estimated downstream influence. Recent work has explored various approaches to select optimal visual instruction tuning datasets. On the other hand, authors in~\citet{chen2024your} score VIT data using an external scoring model trained with the target model. More recently, works such as~\citet{yan2025coido} introduce a lightweight scorer that jointly optimizes importance and diversity using a coupled objective trained on a sampled subset of data. In contrast, MAGIC is entirely training-free and forward-only. Instead of depending on gradient computation, auxiliary scorer optimization, or clustering in continuous feature spaces, it selects samples using intrinsic signals that directly capture multimodal utility, grounding strength, and activation-aware behavioral coverage.

\section{Methodology}
\begin{figure*}[t]
\centering
    \includegraphics[width=0.7\linewidth]{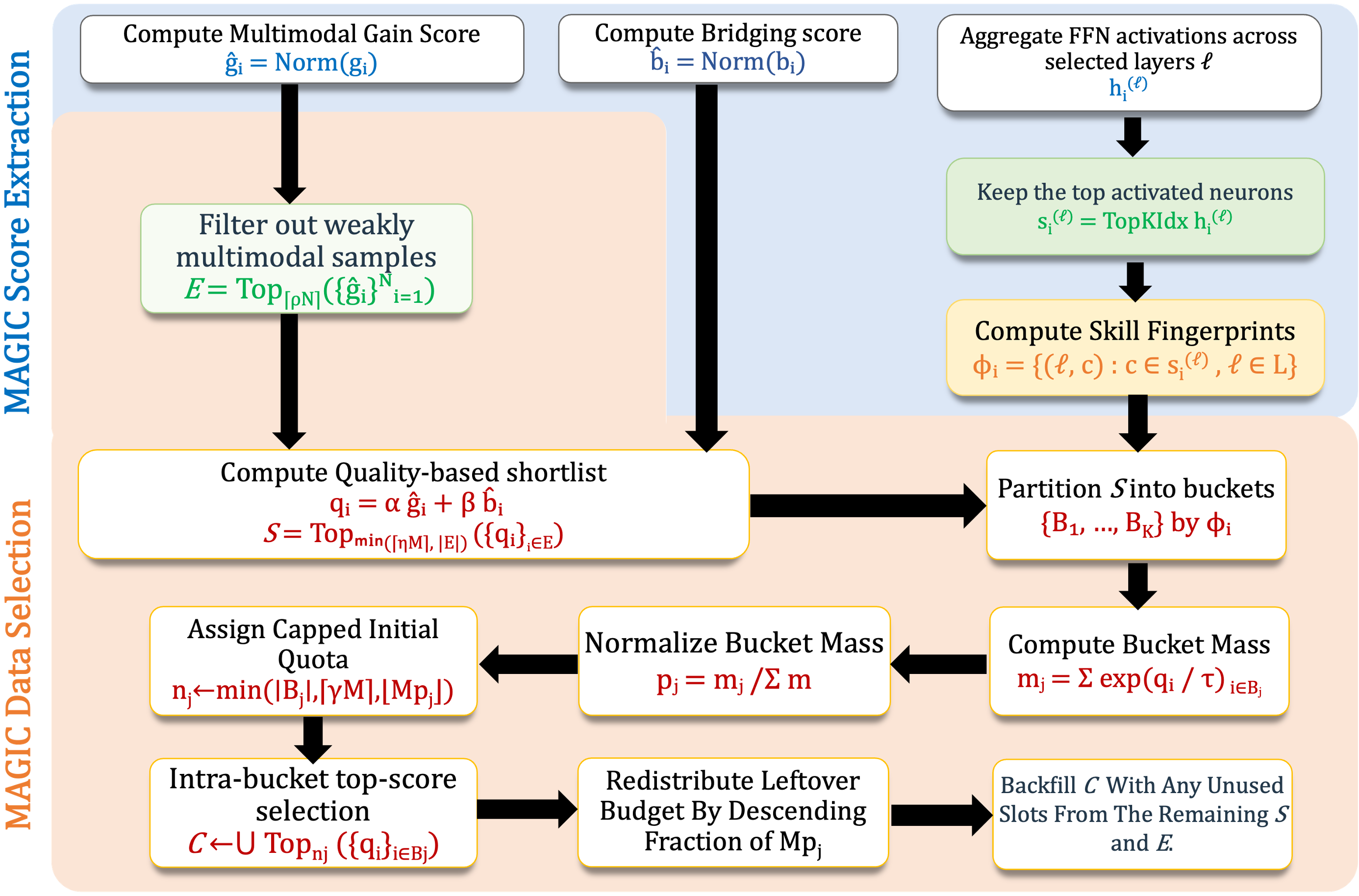} 
    \vspace{-5pt}
    \caption{Overview of MAGIC. The method first extracts three forward-only intrinsic score signals from a pretrained VLM. It then filters weakly multimodal samples, builds a quality-based shortlist, partitions shortlisted samples into skill-signature buckets, and allocates the budget across buckets to construct the final coreset.
    }
    \vspace{-5pt}
    \label{fig:method}
\end{figure*}

In this section, we formalize the MAGIC framework. We begin by introducing the VLM setting and the internal representations accessible from a pretrained model under a forward pass. These quantities form the basis for the three intrinsic signals used by MAGIC, as seen in Fig.~\ref{fig:method}.

\subsection{Problem Formulation}

Let $
\mathcal{D}=\{(I_i,T_i,Y_i)\}_{i=1}^N$
be a visual instruction tuning dataset, where \(I_i\) is an image, \(T_i\) is a textual instruction, and \(Y_i\) is the target response. The goal of multimodal coreset selection is to construct a subset
$\mathcal{C}\subseteq \mathcal{D},$, where, $ |\mathcal{C}|=M \ll N$,
such that finetuning a VLM on \(\mathcal{C}\) retains the utility of finetuning on the full dataset \(\mathcal{D}\) across multiple downstream tasks, whilst reducing visual instruction tuning costs.

\subsection{Preliminaries}

We consider a pretrained LVLM \(\mathcal{M}\) composed of a visual encoder, a modality projector, and a transformer-based language model. For a sample
$x_i=(I_i,T_i,Y_i)$, the image \(I_i\) is encoded into \(N_v\) visual tokens, projected into the language model space, and concatenated with \(N_t\) textual tokens from the instruction-response sequence~\cite{dai2023instructblip, liu2023visual}. Here, we focus on the common setting where visual information is injected into a transformer-based LLM as input tokens. 

At transformer layer \(\ell\), let
\vspace{-5pt}
\begin{equation}
X_i^{(\ell)}=[X_{v,i}^{(\ell)};X_{t,i}^{(\ell)}]\in\mathbb{R}^{(N_v+N_t)\times D}
\vspace{-5pt}
\end{equation}
denote the concatenated hidden states, where \(D\) is the hidden dimension. The multi-head self-attention (MSA) block produces
\vspace{-5pt}
\begin{equation}
Z_i^{(\ell)}=\mathrm{MSA}^{(\ell)}\!\big(\mathrm{LN}^{(\ell)}(X_i^{(\ell)})\big)+X_i^{(\ell)},
\vspace{-5pt}
\end{equation}
where $\mathrm{LN}$ denotes layer normalization, and $Z_i^{(\ell)}=[Z_{v,i}^{(\ell)};Z_{t,i}^{(\ell)}]$ are the output visual and textual features, respectively. Let
\vspace{-5pt}
\begin{equation}
A_i^{(\ell)}\in\mathbb{R}^{n_h\times (N_v+N_t)\times (N_v+N_t)}
\vspace{-5pt}
\end{equation}
denote the corresponding head-wise attention weights induced by \(\mathrm{MSA}^{(\ell)}\), where \(n_h\) is the number of attention heads. We further denote by
\vspace{-5pt}
\begin{equation}
H_i^{(\ell)} \in \mathbb{R}^{N_t \times D_{\mathrm{ff}}}
\vspace{-5pt}
\end{equation}
the token-wise textual FFN activations at layer \(\ell\), where \(D_{\mathrm{ff}}\) is the FFN width.

Under this notation, MAGIC derives three intrinsic forward-pass signals from a pretrained LVLM.

\begin{algorithm*}[t]
\caption{MAGIC Score Extraction}
\label{alg:magic_scores}
\small
\textbf{Input:} dataset $\mathcal{D}=\{(I_i,T_i,Y_i)\}_{i=1}^N$, pretrained LVLM $\mathcal{M}$, selected layers $\mathcal{L}$, per-layer signature sizes $\{k_\ell\}_{\ell\in\mathcal{L}}$\\
\textbf{Output:} multimodal gain $\{g_i\}_{i=1}^N$, bridging relevance $\{b_i\}_{i=1}^N$, skill signatures $\{\phi_i\}_{i=1}^N$
\begin{algorithmic}[1]
\FOR{$i=1,\dots,N$}
    \STATE Compute answer-token set $\mathcal{T}_i^a$ from $Y_i$
    \FOR{each $t\in\mathcal{T}_i^a$}
        \STATE $\Delta_i^{(t)}
        \leftarrow
        \mathrm{CE}(y_{i,t};\, y_{i,<t}, \varnothing, T_i;\theta)
        -
        \mathrm{CE}(y_{i,t};\, y_{i,<t}, I_i, T_i;\theta)$
        \hfill $\textcolor{cyan}{\triangleright}$ \textcolor{cyan}{\mbox{per-token multimodal gain}}
    \ENDFOR
    \STATE $g_i
    \leftarrow
    \frac{1}{|\mathcal{T}_i^{a}|}
    \sum_{t\in\mathcal{T}_i^{a}}
    \Delta_i^{(t)}$
    \hfill $\textcolor{cyan}{\triangleright}$ \textcolor{cyan}{\mbox{multimodal gain}}
    \FOR{each $\ell\in\mathcal{L}$}
        \STATE $A_i^{(\ell)} \leftarrow \mathrm{Attn}^{(\ell)}\!\left(\mathcal{M}(I_i,T_i,Y_i)\right)$
        \hfill $\textcolor{cyan}{\triangleright}$ \textcolor{cyan}{\mbox{head-wise attention tensor}}
        \STATE $\bar{A}_i^{(\ell)} \leftarrow \mathrm{MeanHead}\!\left(A_i^{(\ell)}\right)$
        \hfill $\textcolor{cyan}{\triangleright}$ \textcolor{cyan}{\mbox{average over heads}}
        \STATE $H_i^{(\ell)} \leftarrow \mathrm{FFNAct}^{(\ell)}\!\left(\mathcal{M}(I_i,T_i,Y_i)\right)$
        \hfill $\textcolor{cyan}{\triangleright}$ \textcolor{cyan}{\mbox{token-wise FFN activations}}
        \STATE $h_i^{(\ell)} \leftarrow \frac{1}{|\mathcal{T}_i^a|}\sum_{t\in\mathcal{T}_i^a}H_{i,t}^{(\ell)}$
        \hfill $\textcolor{cyan}{\triangleright}$ \textcolor{cyan}{\mbox{mean answer-token activation}}
        \STATE $s_i^{(\ell)} \leftarrow \mathrm{TopKIdx}_{k_\ell}\!\left(h_i^{(\ell)}\right)$
        \hfill $\textcolor{cyan}{\triangleright}$ \textcolor{cyan}{\mbox{top-$k_\ell$ neurons at layer $\ell$}}
    \ENDFOR
    \STATE $\tilde{A}_{i,tv}^{(\ell)}
=
\frac{\bar{A}_{i,tv}^{(\ell)}}
{\sum_{u=1}^{N_v}\bar{A}_{i,tu}^{(\ell)}}$ \hfill $\textcolor{cyan}{\triangleright}$ \textcolor{cyan}{\mbox{normalized image-token attention}}
    \STATE $e_{i,t}^{(\ell)}
=
-\sum_{v=1}^{N_v}
\tilde{A}_{i,tv}^{(\ell)}
\log \tilde{A}_{i,tv}^{(\ell)}$ \hfill $\textcolor{cyan}{\triangleright}$ \textcolor{cyan}{\mbox{attention entropy}}
    \STATE $b_i
=
\frac{1}{|\mathcal{L}|}
\sum_{\ell\in\mathcal{L}}
\frac{1}{|\mathcal{T}_i^{a}|}
\sum_{t\in\mathcal{T}_i^{a}}
\left[
\left(\sum_{v=1}^{N_v}\bar{A}_{i,tv}^{(\ell)}\right)
\left(
1-\frac{e_{i,t}^{(\ell)}}{\log N_v}
\right)
\right]$
\hfill $\textcolor{cyan}{\triangleright}$ \textcolor{cyan}{\mbox{bridging relevance}}
    \STATE $\phi_i \leftarrow \{(\ell,c): c\in s_i^{(\ell)},\ \ell\in\mathcal{L}\}$
    \hfill $\textcolor{cyan}{\triangleright}$ \textcolor{cyan}{\mbox{multi-layer skill signature}}
\ENDFOR
\STATE \textbf{return} $\{g_i\}_{i=1}^N,\{b_i\}_{i=1}^N,\{\phi_i\}_{i=1}^N$
\end{algorithmic}
\end{algorithm*}

\subsection{MAGIC Score Extraction}
\paragraph{Multimodal Gain.}
To quantify the extent to which a training sample genuinely benefits from multimodal evidence, we compare the model’s answer-token prediction error under full multimodal conditioning and under a unimodal text-only setting ($I_i =\varnothing$). Let \(\mathcal{T}_i^{a}\) denote the set of answer-token positions for sample \(i\). We define the \emph{Multimodal Gain} score as
\vspace{-5pt}
\begin{align}
\vspace{-6pt}
\Delta_i^{(t)}
&=
\mathrm{CE}(y_{i,t};\, y_{i,<t}, \varnothing, T_i;\theta)
\nonumber\\
&\quad-
\mathrm{CE}(y_{i,t};\, y_{i,<t}, I_i, T_i;\theta).
\vspace{-6pt}
\end{align}
\begin{equation}
\vspace{-6pt}
g_i
=
\frac{1}{|\mathcal{T}_i^{a}|}
\sum_{t\in\mathcal{T}_i^{a}}
\Delta_i^{(t)}
\vspace{-5pt}
\end{equation}

where \(\mathrm{CE}(\cdot)\) denotes the per-token cross-entropy, \(I_i\) is the image, \(T_i\) is the textual context, and \(\varnothing\) denotes the absence of visual input. Thus, \(g_i\) quantifies the average reduction in answer-token prediction error attributable to the presence of the image. Samples with large \(g_i\) are those for which visual information materially improves target prediction, indicating high multimodal utility. In contrast, samples with small or near-zero values are largely recoverable from textual context alone and thus provide limited additional supervision for learning image-grounded behavior.

\paragraph{Bridging Relevance.}
Multimodal gain measures whether vision is useful, but not whether the response is strongly grounded in visual evidence. To capture this, we use the attention maps from selected transformer layers \(\mathcal{L}\). For each \(\ell\in\mathcal{L}\), let \(A_i^{(\ell)}\) denote the head-wise attention tensor, and let \(\bar{A}_i^{(\ell)}\) be the average over heads. We consider the submatrix corresponding to answer-token queries $t$ attending to visual-token keys $v$. 
To characterize how this attention mass is distributed within the visual-token block, we define the normalized image-token attention as
\vspace{-12pt}
\begin{equation}
\tilde{A}_{i,tv}^{(\ell)}
=
\frac{\bar{A}_{i,tv}^{(\ell)}}
{\sum_{u=1}^{N_v}\bar{A}_{i,tu}^{(\ell)}}.
\vspace{-8pt}
\end{equation}
This induces a distribution over visual tokens conditioned on answer-token query \(t\). We then define its entropy as $e_{i,t}^{(\ell)}
=
-\sum_{v=1}^{N_v}
\tilde{A}_{i,tv}^{(\ell)}
\log \tilde{A}_{i,tv}^{(\ell)}$. Lower entropy corresponds to attention concentrating on more localized visual evidence, while high entropy indicates diffused attention on the entire image input. 
Hence, we define \emph{Bridging Relevance} as
\vspace{-5pt}
\begin{equation}
\small
b_i
=
\frac{1}{|\mathcal{L}|}
\sum_{\ell\in\mathcal{L}}
\frac{1}{|\mathcal{T}_i^{a}|}
\sum_{t\in\mathcal{T}_i^{a}}
\left[
\left(\sum_{v=1}^{N_v}\bar{A}_{i,tv}^{(\ell)}\right)
\left(
1-\frac{e_{i,t}^{(\ell)}}{\log N_v}
\right)
\right]
\vspace{-5pt}
\end{equation}
\normalsize
Thus, \(b_i\) is high when answer tokens attend strongly to the visual input and concentrate this attention on a small set of evidence-bearing visual tokens.

\paragraph{Skill-Neuron Signature.}
To characterize the latent computation induced by a sample, we extract FFN activations from selected transformer layers, following prior work suggesting that these layers function as key-value memories and that individual neurons can encode interpretable concepts or localized knowledge~\cite{geva2021transformer, geva2022transformer, dai2022knowledge}. Let
$H_i^{(\ell)}$ denote the token-wise textual FFN activations at layer \(\ell\). We aggregate them over the supervised answer-token positions:
\vspace{-7pt}
\begin{equation}
h_i^{(\ell)}
=
\frac{1}{|\mathcal{T}_i^{a}|}
\sum_{t\in\mathcal{T}_i^{a}}
H_{i,t}^{(\ell)} \in \mathbb{R}^{D_{\mathrm{ff}}}.
\vspace{-9pt}
\end{equation}
We then retain the indices of the top-\(k_\ell\) activated neurons,
\vspace{-14pt}
\begin{equation}
s_i^{(\ell)}=\mathrm{TopKIdx}_{k_\ell}\!\left(h_i^{(\ell)}\right),
\vspace{-10pt}
\end{equation}
and form the multi-layer skill signature by layer-neuron pairs as follows:
\vspace{-5pt}
\begin{equation}
\phi_i=\{(\ell,c): c\in s_i^{(\ell)},\ \ell\in\mathcal{L}\}.
\vspace{-5pt}
\end{equation}
This discrete signature serves as a compact behavioral fingerprint of the sample. The resulting triplet \((g_i,b_i,\phi_i)\), as seen in Alg.~\ref{alg:magic_scores}, forms the basis of the subsequent coreset selection procedure. 

\subsection{MAGIC Data Selection}
Given the extracted sample descriptors \((g_i,b_i,\phi_i)\), MAGIC constructs the coreset through quality scoring, shortlist formation, and behavior-aware budget allocation, as shown in Alg.~\ref{alg:magic_select}.

\paragraph{Quality scoring and Eligibility Filtering.}
Since multimodal gain and bridging relevance have different scales, we first apply robust normalization:
\vspace{-10pt}
\begin{equation}
\hat{g}_i=\mathrm{Norm}(g_i), \qquad \hat{b}_i=\mathrm{Norm}(b_i).
\vspace{-3pt}
\end{equation}
We then define a joint quality score $q_i=\alpha \hat{g}_i+\beta \hat{b}_i$, where \(\alpha,\beta\geq 0\) control the relative contribution of multimodal utility and grounding strength. To suppress weakly multimodal examples, we first retain only the top \(\lceil \rho N\rceil\) examples ranked by \(g_i\):
\vspace{-6pt}
\begin{equation}
\mathcal{E}=\mathrm{Top}_{\lceil \rho N\rceil}(\{g_i\}_{i=1}^N).
\vspace{-6pt}
\end{equation}
From this eligible set, we form a quality shortlist
\vspace{-6pt}
\begin{equation}
\mathcal{S}
=
\mathrm{Top}_{\min(\lceil \eta M\rceil,|\mathcal{E}|)}
(\{q_i\}_{i\in\mathcal{E}}),
\vspace{-6pt}
\end{equation}
where \(\eta \ge 1\) is a shortlist expansion factor. This stage ensures that the final selection is drawn from candidates with both strong multimodal utility and high overall quality.

\paragraph{Behavior-Aware Grouping.}
To preserve latent behavioral diversity, we partition the shortlist \(\mathcal{S}\) into discrete buckets according to the skill-neuron signatures \(\phi_i\). Let
\vspace{-6pt}
\begin{equation}
\mathcal{B}=\{B_1,\dots,B_K\}
\vspace{-6pt}
\end{equation}
denote the resulting partition, where samples sharing the same multi-layer signature are assigned to the same bucket. 
Unlike continuous clustering approaches~\cite{yu2025mastering}, this discrete grouping is computationally lightweight and directly captures sample-induced behavioral similarity through shared skill-neuron activation patterns.

\paragraph{Final Selection.}
We allocate the coreset budget across buckets using a temperature-scaled quality mass. For each bucket \(B_j\), we define $m_j=\sum_{i\in B_j}\exp(q_i/\tau)$, and normalize across buckets:
\vspace{-6pt}
\begin{equation}
p_j=\frac{m_j}{\sum_{r=1}^{K}m_r}.
\vspace{-6pt}
\end{equation}
The initial quota for bucket \(B_j\) is then
\vspace{-4pt}
\begin{equation}
n_j=
\min\!\left(
|B_j|,
\left\lceil \gamma M \right\rceil,
\left\lfloor M p_j \right\rfloor
\right),
\vspace{-4pt}
\end{equation}
where \(\gamma\) caps over-allocation to dominant buckets. Any remaining budget
\(
M-\sum_{j=1}^{K}n_j
\)
is redistributed by descending fractional remainder of \(Mp_j\). Finally, we select the top-\(n_j\) samples in each bucket according to \(q_i\):
\vspace{-6pt}
\begin{equation}
\mathcal{C}
=
\bigcup_{j=1}^{K}
\mathrm{Top}_{n_j}(\{q_i\}_{i\in B_j}).
\vspace{-6pt}
\end{equation}
If \(|\mathcal{C}|<M\), we backfill from \(\mathcal{S}\setminus\mathcal{C}\), and then from \(\mathcal{E}\setminus\mathcal{C}\), in descending order of \(q_i\), until the target budget is reached. Here, $\mathcal{A}\setminus\mathcal{B}$ denotes the elements in A that are not already contained in B. Overall, this procedure retains high-quality, strongly grounded, and multimodally useful samples while preserving coverage over distinct latent computation patterns.

\begin{table*}[t]
\centering
\small
\renewcommand{\arraystretch}{1.0}
\resizebox{\textwidth}{!}{%
\begin{tabular}{lcccccccccccc}
\toprule
\rowcolor{gray!20}
\textbf{Method} & \textbf{MME} & \textbf{SQA-I} & \textbf{POPE} & \textbf{VQAv2} & \textbf{LLaVA-W Bench} & \textbf{TextVQA} & \textbf{MMBench en} & \textbf{MMBench cn} & \textbf{GQA} & \textbf{VizWiz} & \textbf{MM-Vet} & \textbf{Rel. (\%)} \\
\midrule
Full              & 1476.9 & 68.4 & 86.4 & 79.1 & 67.9 & 58.2 & 66.1 & 58.9 & 63.0 & 47.8 & 30.9 & 100.0 \\ \midrule
Random            & 1483.0 & 68.5 & 84.7 & 75.7 & 65.0 & 55.3 & 62.2 & 54.8 & 58.9 & 44.3 & 29.5 & 95.8 \\
CLIP-Score   & 1331.6 & 65.0 & 85.3 & 73.4 & 66.2 & 54.7 & 55.2 & 52.0 & 51.4 & 43.0 & - & 91.2 \\
EL2N         & 1439.5 & 65.5 & 84.3 & 76.2 & 64.9 & 53.0 & 53.2 & 47.4 & 58.7 & 43.7 & 21.1 & 89.8 \\
Perplexity   & 1341.4 & 65.1 & 82.6 & 75.8 & 68.3 & 52.8 & 52.0 & 45.8 & 57.0 & 47.8 & - & 91.6 \\
SemDeDup     & 1376.9 & 65.8 & 84.7 & 74.2 & \underline{70.0} & \underline{55.5} & 52.2 & 48.5 & 54.5 & 46.9 & - & 92.6 \\
D2-Pruning   & 1391.2 & 69.3 & 85.7 & 73.0 & 63.9 & 51.8 & \textbf{65.7} & \textbf{57.6} & 58.4 & 41.9 & - & 94.8 \\
Self-Sep     & 1335.9 & 67.8 & 83.5 & 74.9 & 63.3 & 49.3 & 61.4 & 53.8 & 59.5 & 46.0 & - & 93.4 \\
Self-Filter  & 1306.2 & 61.4 & 83.8 & 73.7 & 64.9 & 52.9 & 48.8 & 45.3 & 58.3 & \textbf{53.2} & 26.6 & 90.5 \\
COINCIDE     & \underline{1495.6} & 69.2 & 86.1 & 76.5 & 67.3 & \underline{55.6} & 63.1 & 54.5 & 59.8 & 46.8 & - & 97.4 \\
RDS          & 1093.8 & 68.0 & 86.3 & 75.1 & 63.7 & 54.9 & 61.2 & 52.7 & 57.9 & 48.6 & - & 93.2 \\
DataTailor   & 1476.1 & - & 85.3 & - & - & - & - & - & 49.5 & 31.8 & - & 85.9 \\
COIDO        & 1450.2 & 69.4 & 85.4 &\textbf{77.2}  & \textbf{70.1}& 55.6 & 63.8& 56.7 &  \underline{60.4} & 47.1& - & 98.2 \\
ICONS        & 1485.7 & \underline{70.8} & \textbf{87.5} & 76.3 & 66.1 & \underline{55.6} & 63.1 & 55.8 & \textbf{60.7} & 50.1 & \underline{29.7} & \underline{98.4} \\
\rowcolor{cyan!10}
MAGIC (ours)         & \textbf{1657.9} & \textbf{72.3} & \underline{87.1} & \underline{76.8} & 69.0 & \textbf{56.4} & \underline{65.1} & \underline{56.8} & \underline{60.4} & \underline{50.2} & \textbf{29.8} & \textbf{100.3} \\
\bottomrule
\end{tabular}}
\vspace{-5pt}
\caption{\textbf{Performance comparison of different data selection approaches when trained on 20\% of the LLaVA-665K dataset.} The best and second-best results for each benchmark are shown in \textbf{bold} and \underline{underlined}, respectively. Our method MAGIC achieves the highest overall Rel. (100.3\%), consistently outperforming existing approaches including ICONS (98.4\%) and COIDO (98.2\%).}
\vspace{-2pt}
\label{tab:llava665k_selection}
\end{table*}

\begin{table*}[t]
\centering
\small
\setlength{\tabcolsep}{4pt}
\renewcommand{\arraystretch}{}
\resizebox{\textwidth}{!}{%
\begin{tabular}{l c l  c c c c c c c  cc c}
\toprule
\rowcolor{gray!20}
\textbf{Dataset} & \textbf{\#Data} & \textbf{Method} & \textbf{VQAv2} & \textbf{GQA} 
& \textbf{SQA-I} & \textbf{TextVQA} & \textbf{POPE} & \textbf{MME} & \shortstack{\textbf{MMBench}\\\textbf{en}} & \shortstack{\textbf{MMBench}\\\textbf{cn}} & \shortstack{\textbf{LLaVA-W Bench}\\\textbf{Bench}} & \textbf{Rel. (\%)} \\
\midrule

\multirow{3}{*}{Vision-Flan-186K}
& 186k   & Full-Finetune           & 68.0 & 49.2 & 
60.8 & 50.4 & 83.4 & 1263.2 & 52.6 & 45.9&63.3 & 100.0 \\ 
& $\sim$37k & Random        & 64.1 & 45.8 
& 58.7 & 45.3 & 82.9 & 1079.8 & 46.5 & 39.6 &58.7& 91.8 \\
\rowcolor{cyan!10}
& $\sim$37k & \textbf{MAGIC (ours)} & 67.5 & 50.7 & 65.6 &  45.7& 84.8 & 1351.6  & 52.0& 46.5&  63.2&  101.6\\
\bottomrule
\end{tabular}%
}
\vspace{-5pt}
\caption{\textbf{Selection results on Vision-Flan-186K.} Performance comparison of different data selection approaches when trained on 20\% of the Vision-Flan-186K. MAGIC achieves strong performance (101.6\%) while using only 20\% of the training data, significantly outperforming random selection and full performance.}
\vspace{-2pt}
\label{tab:visionflan_cambrian_selection}
\end{table*}
\section{Evaluation}
\subsection{Experimental Setup}

\textbf{Datasets.}
We conduct 20\% coreset selection experiments on two visual instruction tuning datasets: LLaVA-665K~\cite{liu2024improved} and Vision-Flan-186K~\cite{xu2024vision}. LLaVA-665K contains approximately 665K multimodal instruction samples collected from diverse vision-language tasks, while Vision-Flan-186K comprises 191 vision-language tasks with roughly 1K expert-annotated instruction examples per task, totaling 186K samples.

\textbf{Models for Selection and Target Finetuning.} Following prior work~\cite{lee2024concept}, we use TinyLLaVA-2B~\cite{zhou2024tinyllava} as our reference model for efficient data selection, from which MAGIC extracts its scores. Unless otherwise specified, our primary target model is LLaVA-1.5-7B~\cite{liu2024improved}. To evaluate transferability of the selected coreset to larger backbones, we additionally evaluate performance on LLaVA-1.5-13B. Following the standard LLaVA-1.5 finetuning recipe, we finetune using LoRA~\cite{hu2022lora} for one epoch with the official hyperparameter settings, using 4 NVIDIA A100 GPUs. The joint quality score uses weights $(\alpha, \beta) = (0.5, 0.5)$ for MG and BR, respectively, and we extracted scores from $4$ uniformly spaced layers from TinyLLaVA-2B, inspired by prior work~\cite{lee2024concept}. More details on experimental settings is in App. Sec.~\ref{app:expdetails}.

\textbf{Evaluation Benchmarks.}
We evaluate on a diverse benchmark suite covering key VLM capabilities, including multiple-choice understanding, visual question answering, text understanding in images, scientific reasoning, open-ended generation, and factual consistency. Concretely, our evaluation suite includes MMBench~\cite{liu2024mmbench}, MME~\cite{fu2023mme}, VQAv2~\cite{goyal2017making}, GQA~\cite{hudson2019gqa}, VizWiz~\cite{gurari2018vizwiz}, TextVQA~\cite{singh2019towards}, ScienceQA~\cite{saikh2022scienceqa}, LLaVA-W Bench~\cite{lu2022learn}, MM-Vet~\cite{yu2023mm}, and POPE~\cite{li2023evaluating}. 

\textbf{Baselines.}
We compare MAGIC against several strong baselines spanning different data selection paradigms. These include random selection, CLIP-Score~\cite{radford2021learning} for measuring image-text alignment, EL2N~\cite{paul2021deep} based on prediction difficulty, and Perplexity~\cite{marion2023less} using language modeling uncertainty. We also compare against SemDeDup~\cite{abbas2023semdedup} for semantic deduplication, D2-Pruning~\cite{maharana2023d2} for distribution-aware data pruning, Self-Sup~\cite{sorscher2022beyond}, which leverages self-supervised signals, and Self-Filter~\cite{chen2024your}, which uses self-filtering to remove less informative samples. Additional baselines include methods designed specifically for vision-language coresets, such as COINCIDE~\cite{lee2024concept}, DataTailor~\cite{yu2025mastering}, COIDO~\cite{yan2025coido}, ICONS~\cite{wu2024icons}, and representation-based selection (RDS)~\cite{xia2024less, ivison2025large}. 


\begin{table*}[t]
\centering
\small
\setlength{\tabcolsep}{4pt}
\renewcommand{\arraystretch}{}
\resizebox{0.87\textwidth}{!}{%
\begin{tabular}{l|c c c c c c c c c|c}
\hline
\rowcolor{gray!15}
\textbf{Method} & \textbf{VQAv2} & \textbf{GQA} & \textbf{SQA-I} & \textbf{TextVQA} & \textbf{POPE} & \textbf{MME} & \shortstack{\textbf{MMBench}\\\textbf{en}\;\;\;\;\textbf{cn}} & \shortstack{\textbf{LLaVA-}\\\textbf{W Bench}} & \textbf{MM-Vet} & \textbf{Rel. (\%)} \\
\hline
Full-Finetune   & 80.0 & 63.3 & 71.2 & 60.2 & 86.7 & 1541.7 & 68.5 \;\;\; 61.5 & 69.5 & 36.1 & 100.0 \\
\hline
Random          & 76.7 & 60.5 & 68.8 & 57.7 & 84.8 & 1484.9 & 62.8 \;\;\; 55.2 & 68.6 & \underline{35.5} & 95.7 \\
CLIP-Score      & 75.3 & 52.6 & 69.7 & 57.3 & 85.4 & 1426.3 & 60.4 \;\;\; 54.0 & 68.1 & - & 92.8 \\
EL2N            & 77.2 & 59.6 & 69.9 & 56.1 & 84.1 & \underline{1531.0} & 59.3 \;\;\; 52.3 & 65.8 & - & 93.8 \\
Perplexity      & 77.0 & 58.5 & 68.7 & 54.8 & 83.1 & 1508.8 & 57.5 \;\;\; 50.3 & \underline{68.7} & - & 92.7 \\
SemDedup        & 75.6 & 57.5 & 70.5 & 57.7 & 85.3 & 1397.6 & 59.0 \;\;\; 51.1 & 68.7 & - & 93.0 \\
D2-Pruning      & 73.9 & 60.5 & \underline{70.4} & 55.2 & 84.9 & 1463.0 & \underline{67.3} \;\;\; \underline{59.9} & 66.5 & - & 95.9 \\
Self-Sup        & 76.3 & \textbf{60.5} & 70.2 & 52.7 & 85.4 & 1463.8 & 63.7 \;\;\; 57.6 & 64.9 & - & 94.5 \\
Self-Filter     & 75.0 & 59.8 & 69.5 & 55.8 & 84.5 & 1446.9 & 58.8 \;\;\; 51.8 & 69.1 & - & 93.3 \\
COINCIDE        & \textbf{77.8} & \underline{60.4} & 70.0 & \underline{58.6} & \textbf{87.1} & 1516.8 & 64.0 \;\;\; 57.7 & 67.4 & - & \underline{96.8} \\
\rowcolor{cyan!10}
MAGIC (Ours)    & 78.9 & \underline{60.4} & \textbf{73.5} & \textbf{58.8} & \underline{86.4} & \textbf{1551.5} & \textbf{67.4} \;\;\; \textbf{61.3} & \textbf{69.0} & \textbf{36.3} & \textbf{99.3} \\
\hline
\end{tabular}%
}\vspace{-5pt}
\caption{\textbf{Transferring to the larger target model.} We validate if the coresets selected from TinyLLaVA-2B are transferable to LLaVA-1.5-13B finetuning, and measure performance on various multimodal benchmarks.}
\label{tab:transfer_larger_model}
\end{table*}

\begin{table*}[t]
\centering
\small
\setlength{\tabcolsep}{4pt}
\renewcommand{\arraystretch}{}
\resizebox{0.87\textwidth}{!}{%
\begin{tabular}{l|c c c c c c c c|c}
\hline
\rowcolor{gray!15}
\textbf{} & \textbf{AI2D} & \textbf{ChartQA} & \textbf{DocVQA} & \textbf{InfoVQA} & \textbf{MMVet} & \textbf{Naturalbench} & \textbf{RealworldQA} & \textbf{CMMMU} & \textbf{Rel. (\%)} \\
\hline
Full-Finetune                & 55.4 & 17.5 & 28.9 & 26.5 & 31.1 & 12.4 & 52.4 & 22.1 & 100.0 \\
\hline
Random             & 50.2 & 15.1 & 25.2 & 24.3 & 29.6 & 11.1 & 49.8 & 21.9 & 91.8 \\
CLIP-Score         & 52.0 & 16.4 & 27.1 & 24.9 & 29.2 & 11.6 & 49.3 & 20.8 & 93.9 \\
EL2N               & 52.1 & 17.9 & 26.9 & 26.4 & 29.6 & 12.5 & 51.9 & 21.7 & 97.8 \\
Perplexity         & 53.0 & 16.4 & 27.1 & 27.1 & 29.4 & 11.4 & 50.3 & 23.9 & 97.0 \\
SemDeDup           & 53.2 & 16.4 & 26.4 & 26.6 & 29.1 & 13.2 & 51.1 & 22.8 & 97.8 \\
D2-Pruning         & 52.1 & 17.3 & 26.5 & 24.7 & 33.1 & 11.3 & 52.0 & 21.9 & 96.7 \\
Self-Sup           & 52.7 & 16.6 & 27.4 & 25.2 & 29.6 & 11.8 & 50.1 & 21.0 & 95.0 \\
Self-Filter        & 52.1 & 16.0 & 27.1 & 25.0 & 29.3 & 11.5 & 49.5 & 20.6 & 94.0 \\
COINCIDE           & 53.7 & 17.0 & 27.9 & 26.0 & 30.2 & 12.1 & 50.9 & 21.4 & 97.2 \\
RDS                & 53.0 & 16.8 & 27.5 & 25.4 & 29.8 & 12.0 & 50.3 & 21.3 & 95.6 \\
ICONS   & 53.9 & 17.1 & 27.9 & 27.5 & 29.7 & 12.8 & 55.0 & 25.2 & 101.6 \\
\rowcolor{cyan!10}
\rowcolor{cyan!10}
MAGIC (ours)   & 54.9 & 17.1 & 27.2 & 26.9 & 29.8 & 13.2 & 54.5 & 25.3 & 101.7 \\
\rowcolor{gray!10}
\rowcolor{gray!10}
Per-task Rel. (\%) & 99.1 & 97.7 & 94.1 & 101.5 & 95.8 & 106.5 & 104.0 & 114.5 & - \\
\hline
\end{tabular}%
}
\vspace{-5pt}
\caption{\textbf{Detailed results of unseen task generalization.} Performance comparison on unseen benchmarks when trained on selected 20\% subsets. Notably, we observe improvements on InfoVQA (101.5\%), RealWorldQA (104.0\%), and CMMMU (114.5\%), highlighting strong generalization to unseen tasks.}
\vspace{-5pt}
\label{tab:unseen_task_generalization}
\end{table*}

\begin{table*}[t]
\centering
\small
\setlength{\tabcolsep}{5pt}
\renewcommand{\arraystretch}{1.12}
\resizebox{0.9\textwidth}{!}{%
\begin{tabular}{lccccc}
\toprule
\rowcolor{gray!15}
\textbf{Method} & \textbf{End-to-End Cost (GPU-hr)} $\downarrow$ & \textbf{Selection Cost (GPU-hr)} $\downarrow$ & \textbf{Finetuning Cost (GPU-hr)} $\downarrow$ & \textbf{External API} & \textbf{Rel. Perf. (\%)} $\uparrow$ \\
\midrule
Full-Finetune               & 76.0                & --   & 68.0 & No  & 100.0 \\
COIDO               & 35.0     & 25.0 & 10.0 & Yes & 98.2 \\
Self-Filter         & 81.0                & 71.0 & 10.0 & No  & 90.5 \\
ICONS               & 66.0                & 50.0 & 10.0 & No  & 98.4 \\
DataTailor          & 33.5                & 23.5 & 10.0 & No  & 85.9 \\
COINCIDE            & 66.5                & 55.5 & 10.0 & No  & 97.4 \\
\rowcolor{cyan!10}
\textbf{MAGIC (Ours)} & \textbf{20.0} & \textbf{10.0} & \textbf{10.0} & \textbf{No} & \textbf{100.3} \\
\bottomrule
\end{tabular}}
\vspace{-7pt}
\caption{\textbf{End-to-end efficiency comparison.} MAGIC achieves the best trade-off between compute cost and performance, matching or exceeding full-data performance while requiring the lowest overall compute.}
\vspace{-5pt}
\label{tab:magic_efficiency}
\end{table*}

\begin{table}[t]
\centering
\small
\setlength{\tabcolsep}{5pt}
\renewcommand{\arraystretch}{1.12}
\resizebox{0.8\columnwidth}{!}{%
\begin{tabular}{lccc}
\toprule

\rowcolor{gray!15}
\textbf{Variant} & \textbf{Rel. (\%)} $\uparrow$ & \textbf{POPE} $\uparrow$ & \textbf{GQA} $\uparrow$ \\
\midrule
\rowcolor{cyan!10}
MAGIC (full)      & \textbf{100.0} & \textbf{87.0}  & \textbf{59.9} \\
w/o BR Score  & 98.4          & 86.7          & 58.3 \\
w/o  MG Score    & 98.1          & 86.7           & 58.1 \\
w/o SN Score  & 98.3          & 86.8          & 57.6 \\
\bottomrule
\end{tabular}%
}\vspace{-5pt}
\caption{\textbf{Ablation study of MAGIC.} Removing any component degrades performance.}
\vspace{-5pt}
\label{tab:magic_ablation}
\end{table}

\begin{table}[t]
\centering
\small
\setlength{\tabcolsep}{5pt}
\renewcommand{\arraystretch}{}
\resizebox{0.8\columnwidth}{!}{%
\begin{tabular}{lccc}
\toprule
\rowcolor{gray!15}
\textbf{Hyperparameter} & \textbf{Setting} & \textbf{SQA} $\uparrow$ & \textbf{POPE} $\uparrow$\\
\midrule
MG/BR weights & (1.0, 0.0) & 69.8 &86.7\\
              & (0.7, 0.3) & 71.8 &87.0\\
             \rowcolor{cyan!10} & (0.5, 0.5) & 72.3& 87.1\\
              & (0.3, 0.7) & 70.9&86.9\\
              & (0.0, 1.0) & 71.2&86.8\\
\midrule
top-$(k_\ell)$ & [1,1,1,1]   & 71.8 &86.8\\
& [1,1,2,2]   &  71.1&86.9\\
                              \rowcolor{cyan!10}& [1,1,2,3]   &  
                              73.5&87.0\\
                              & [2,3,4,6]   &  70.2&85.5\\

\bottomrule
\end{tabular}%
}\vspace{-5pt}
\caption{\textbf{Hyperparameter sensitivity of MAGIC.} 
}
\vspace{-5pt}
\label{tab:magic_hparam}
\end{table}

\subsection{Results and Discussion}
\paragraph{Main results on LLaVA-665K.}
Table~\ref{tab:llava665k_selection} summarizes coreset selection results on LLaVA-665K. MAGIC achieves the best overall relative score of 100.3\%, outperforming all baselines, including ICONS (98.4\%) and COIDO (98.2\%). It also delivers the strongest results on several benchmarks, including MME, SQA-I, TextVQA, and MM-Vet, while remaining highly competitive across the remaining evaluation suite. Notably, MAGIC slightly surpasses full-data finetuning in aggregate relative performance despite using only 20\% of the training data, suggesting that it effectively removes redundant or weakly informative samples while retaining high-value multimodal supervision. Overall, these results show that MAGIC provides a superior coreset construction strategy among both score-based and model-involved baselines. We further visualize top selected and bottom rejected samples in Fig.~\ref{fig:sample} to show the data qualitative preference induced by MAGIC, highlighting that the method favors samples with stronger multimodal utility and grounding while rejecting samples that are less visually informative or behaviorally redundant.

\textbf{Main results on Vision-Flan.}
Table~\ref{tab:visionflan_cambrian_selection} reports coreset selection results on Vision-Flan-186K under a 20\% budget. MAGIC achieves a relative score of 101.6\%, outperforming random selection by a large margin (+9.8 points), even exceeding full-data finetuning. These gains are consistent across several benchmarks, including GQA, SQA-I, POPE, MME, and MMBench-cn, indicating that the selected subset preserves not only overall utility but also strong cross-task generalization. This result suggests that the combination of multimodal utility, grounding strength, and behavioral coverage can identify compact training subsets that match or exceed the effectiveness of the full dataset.

\textbf{Transfer to a Larger Target Model.}
To test whether the selected coresets transfer when the target model has a much larger backbone than the reference model, we finetune LLaVA-1.5-13B on the 20\% subsets selected from TinyLLaVA-2B. As shown in Table~\ref{tab:transfer_larger_model}, MAGIC achieves the strongest overall transfer performance, reaching 99.3\% relative performance to full-finetuning. 
MAGIC attains the best or second-best performance across all benchmarks, showing that the proposed criteria identify samples whose value is not tied to a single model scale and that MAGIC captures broadly useful multimodal training signals rather than overfitting to the inductive biases of the reference model.

\textbf{Generalization to Unseen Tasks.}
We further evaluate whether the selected subsets support generalization beyond the benchmarks seen during coreset construction. More details on these datasets are provided in App. Sec.~\ref{app sec:unseen}. Tab.~\ref{tab:unseen_task_generalization} shows that MAGIC achieves the best overall unseen-task relative score (101.7\%), surpassing prior baselines. In particular, MAGIC yields strong relative improvements on InfoVQA (101.5\%), NaturalBench (106.5\%), RealWorldQA (104.0\%), and CMMMU (114.5\%). While performance on a few tasks remains slightly below full-data finetuning, the overall pattern suggests that preserving activation-level behavioral diversity improves robustness to task shift. This supports our central hypothesis that coreset construction should account not only for sample quality, but also for diversity in the latent computations induced by the selected data.

\section{Ablations}
\textbf{Cost of Coreset Selection}
Table~\ref{tab:magic_efficiency} compares end-to-end efficiency across data selection methods. MAGIC achieves the strongest compute-performance trade-off, amongst all baselines. At the same time, MAGIC attains the highest relative performance (100.3\%). Notably, MAGIC also avoids external API usage, unlike COIDO. These results highlight the scalability and effectiveness of our proposed design. 

\textbf{Component Analysis.}
Table~\ref{tab:magic_ablation} evaluates the contribution of each component in MAGIC. Removing either the BR signal, the multimodal-gain eligibility filter, or the SN-based behavioral bucketing consistently degrades performance. 
These results confirm that the three components are complementary: multimodal gain filters out weakly visual examples, bridging relevance favors strongly grounded samples, and skill-signature bucketing preserves coverage over distinct behavioral modes.

\textbf{Hyperparameter Sensitivity.}
Table~\ref{tab:magic_hparam} evaluates sensitivity to the MG/BR weighting and the per-layer signature size \(k_\ell\), which specifies how many top-activated neurons are retained from each selected layer. A balanced ($\alpha,\beta$) weighting works best: \((0.5,0.5)\) achieves the strongest performance, indicating that multimodal utility and grounding strength are most effective when combined. For the signature design, intermediate granularity performs best: \([1,1,2,3]\), which retains 1, 1, 2, and 3 neurons from the four selected layers, yields the strongest overall score, which aligns with the experimental settings MAGIC uses. Smaller signatures under-represent latent behavior, while larger ones make buckets overly specific and less robust by fragmenting samples into overly fine groups. 

\section{Conclusion}
We presented MAGIC, a training-free and forward-only coreset selection method for visual instruction tuning. MAGIC combines three intrinsic signals from a pretrained VLM -- Multimodal Gain, Bridging Relevance, and Skill-Neuron Signatures -- to jointly model multimodal utility, grounding strength, and activation-level behavioral diversity. Comprehensive experiments on the LLaVA-1.5 and Vision-Flan datasets demonstrate that MAGIC consistently achieves stronger or comparable performance to full-data finetuning while using only 20\% of the data, with the lowest data selection cost, showcasing its effectiveness and efficiency. These results show that high-value multimodal coresets should not be defined solely by scalar difficulty or redundancy, but also by the diversity of latent computations they elicit.

\section{Limitations}
MAGIC has a few limitations. First, like most VLM coreset selection methods, our score extraction depends on a pretrained reference LVLM, so selection quality may vary with the choice of reference model. That said, our transfer results suggest that the extracted signals remain useful beyond the exact model used for scoring. Second, MAGIC requires dataset-wide forward passes to extract selection signals. However, this cost is very modest compared to gradient-based or auxiliary-training methods, since MAGIC avoids backpropagation, scorer training, and iterative optimization. In practice, this yields a favorable efficiency-performance trade-off: MAGIC remains simple, training-free, and consistently effective across datasets and transfer settings.

\section{Ethical Considerations and Broader Impact}

MAGIC is designed to improve the efficiency of multimodal instruction tuning by selecting compact yet behaviorally representative coresets for large vision-language models. By reducing the amount of training data required for finetuning, MAGIC can lower computational cost, energy consumption, and environmental impact, enabling more sustainable multimodal model development.

Because MAGIC operates on large-scale web-derived multimodal datasets, the selected subsets may still reflect demographic biases, unsafe content, or skewed visual-textual associations present in the original data. Although our bucket-aware balancing strategy improves coverage across diverse latent behaviors and reduces over-concentration on dominant modes, it does not explicitly enforce fairness or safety constraints. Nevertheless, the structured and diversity-aware design of MAGIC provides a promising foundation for incorporating fairness-aware or safety-aware objectives, which are beyond the scope of this work, into future coreset selection methods.

Finally, while more efficient coreset selection can broaden access to multimodal model training, responsible deployment remains important. We encourage practitioners to combine methods such as MAGIC with dataset auditing, safety filtering, and appropriate governance practices to support reliable and responsible use of large vision-language models.

\section{AI Writing Statement}
This paper utilized AI assistance for language polishing of the manuscript, including vocabulary correction and spell checking.
\bibliography{custom}

\newpage
\appendix

\section{Appendix}
\label{sec:appendix}

This appendix provides supplementary details for the main paper. We introduce the task of visual instruction-tuning, include implementation and experimental details, benchmark descriptions and task statistics, the full MAGIC data selection algorithm, additional coverage analysis in the MAGIC feature space, and qualitative examples of selected and rejected samples. These materials complement the main results by offering further detail on the methodology and additional evidence for the effectiveness of MAGIC.

\subsection{Visual Instruction Tuning} 
Recent vision-language models (VLMs), such as Flamingo~\cite{alayrac2022flamingo}, LLaVA~\cite{liu2024improved}, BLIP-2~\cite{li2023blip}, and Cambrian~\cite{tong2024cambrian}, have substantially advanced multimodal intelligence by coupling visual perception with large language reasoning. Their downstream effectiveness relies heavily on \emph{visual instruction tuning} (VIT)~\cite{liu2023visual}, a post-pretraining adaptation stage that finetunes the model on image-grounded instruction-response tuples. By exposing the model to diverse multimodal task formulations under natural language supervision, VIT improves instruction adherence, strengthens cross-modal alignment, and enables the model to operate as a general-purpose multimodal assistant~\cite{liu2023visual}. In parallel with the expansion of multimodal application domains, VIT datasets have grown to massive scale, now routinely comprising hundreds of thousands to millions of annotated instruction examples~\cite{liu2024improved, tong2024cambrian}.

\subsection{Datasets \& model}
For our experiments, we use the LLaVA-v1.5 checkpoint after Stage~1 pre-training for feature alignment, following the original LLaVA training pipeline, with the 7B model size and LLaVA-665K setting unless otherwise specified. Concretely, we use the \texttt{checkpoint}\footnote{\url{https://huggingface.co/liuhaotian/llava-v1.5-mlp2x-336px-pretrain-vicuna-7b-v1.5}}, which represents the model after projector training and before any Stage~2 visual instruction tuning. Therefore, the checkpoint has no prior exposure to the LLaVA-665K visual instruction tuning data before data selection. 

\subsection{Details on Experimental Settings}
\label{app:expdetails}

Unless otherwise specified, all coreset selection methods use a fixed selection budget of \(20\%\) of the original training set. For LLaVA-665K, this corresponds to \(133{,}059\) selected samples, while for Vision-Flan-186K it corresponds to approximately \(37\)K samples. For all experiments, we use TinyLLaVA-2B as the reference model for score extraction and LLaVA-1.5-7B as the primary target model for finetuning; transfer experiments further evaluate the selected coresets on LLaVA-1.5-13B.

For score extraction in MAGIC, we use the pretrained LLaVA-v1.5 checkpoint after Stage~1 feature alignment, before any visual instruction tuning, so that the reference model has not previously seen the downstream VIT training data. We extract both Bridging Relevance and Skill-Neuron signals from the same four uniformly spaced transformer layers,
\[
\mathcal{L}_{\mathrm{BR}}=\mathcal{L}_{\mathrm{SN}}=\{8,12,16,20\},
\]
taking inspiration from prior work~\cite{lee2024concept}. We choose these layers to capture a broad spectrum of multimodal computations across early, middle, and late stages of the model, while keeping feature extraction efficient. Earlier layers tend to reflect lower-level alignment and local grounding signals, middle layers capture richer cross-modal interaction patterns, and higher layers encode more task-specific semantic reasoning. Using multiple layers distributed from lower to upper depths therefore provides a more stable view of both answer-to-vision grounding and latent behavioral activation than relying on a single layer alone. At the same time, restricting extraction to four layers keeps the procedure efficient and avoids the redundancy and overhead of using the entire network. Score extraction is performed with a per-device batch size of \(2\), and the raw SN extraction retains the top \(64\) activated neurons per selected layer before the later signature compression step used in data selection.

For data selection, we use a target coreset size of \(133{,}059\), corresponding to a \(20\%\) subset of LLaVA-665K. We first apply a multimodal-gain eligibility filter with keep ratio \(\rho=0.6\), then form a quality shortlist with factor \(\eta=2.0\). Unless otherwise specified, the joint quality score uses weights \((\alpha,\beta)=(0.5,0.5)\) for MG and BR, respectively. For skill-signature bucketing, we use the per-layer signature configuration
\[
[k_{\ell}]_{\ell\in\mathcal{L}} = [1,1,2,3],
\]
meaning that we retain 1, 1, 2, and 3 neurons from the four selected layers, respectively. Bucket allocation uses temperature \(\tau=0.2\) and a maximum bucket fraction of \(\gamma=0.05\). We then perform bucket-wise top-score selection followed by quota correction and backfilling as described in Algorithm~\ref{alg:magic_select}.


For target-model finetuning, we follow the official LLaVA-1.5 recipe and train with LoRA for one epoch using the standard hyperparameter settings. All experiments are conducted on 4 NVIDIA A100 GPUs for a single run.

\subsection{Details on Evaluation Benchmarks.} We evaluate on a diverse collection of multimodal benchmarks that capture complementary capabilities of VLMs, including multiple-choice reasoning, visual question answering, text understanding in images, scientific reasoning, open-ended generation, and factual consistency. Specifically, VQAv2~\cite{goyal2017making} measures open-ended question answering over natural images, while GQA~\cite{hudson2019gqa} emphasizes compositional reasoning over object attributes and relations. VizWiz~\cite{gurari2018vizwiz} evaluates robustness on real-world images captured by visually impaired users, and TextVQA~\cite{singh2019towards} focuses on reading and reasoning over scene text. ScienceQA~\cite{saikh2022scienceqa} assesses science-oriented reasoning grounded in visual content. POPE~\cite{li2023evaluating} is used to quantify object hallucination, whereas MME~\cite{fu2023mme} evaluates perception and cognition through a collection of binary-choice subtasks. MMBench~\cite{liu2024mmbench} provides a broad testbed covering abilities such as recognition, OCR, and relational reasoning. In addition, LLaVA-W Bench~\cite{lu2022learn} evaluates open-ended visual instruction-following behavior, and MM-Vet~\cite{yu2023mm} measures a wide range of vision-language skills, including recognition, OCR, knowledge, spatial reasoning, and mathematical understanding. More details on the task statistics are shown in Tab.~\ref{tab:target_task_statistics}.

\begin{table}[t]
\centering
\small
\setlength{\tabcolsep}{3pt}
\renewcommand{\arraystretch}{1.08}
\resizebox{\columnwidth}{!}{%
\begin{tabular}{l|c c c|c c}
\toprule
\textbf{Task} & \textbf{MME} & \textbf{POPE} & \textbf{SQA-I} & \multicolumn{2}{c}{\textbf{MMBench}} \\
              &              &               &                & \textbf{en}      & \textbf{cn}      \\
\midrule
$|\mathcal{D}_{\mathrm{val}}|$  & 986   & 500   & 424   & 1,164 & 1,164 \\
$|\mathcal{D}_{\mathrm{test}}|$ & 2,374 & 8,910 & 4,241 & 1,784 & 1,784 \\
\textbf{Task Type}              & Y/N   & Y/N   & MCQ   & MCQ   & MCQ   \\
\midrule
\textbf{Task} & \textbf{VQAv2} & \textbf{GQA} & \textbf{VizWiz} & \textbf{TextVQA} & \textbf{LLaVA-W} \\
\midrule
$|\mathcal{D}_{\mathrm{val}}|$  & 1,000  & 398    & 800   & 84    & 84 \\
$|\mathcal{D}_{\mathrm{test}}|$ & 36,807 & 12,578 & 8,000 & 5,000 & 84 \\
\textbf{Task Type}              & VQA    & VQA    & VQA   & VQA   & VQA \\
\bottomrule
\end{tabular}%
}\vspace{-5pt}
\caption{\textbf{Statistics of Target Tasks.} Our target tasks include diverse benchmarks and answer formats, covering different vision-language capabilities. Task types include Multiple-Choice Questions (MCQ), Visual Question Answering (VQA), and Yes/No Questions (Y/N).}
\label{tab:target_task_statistics}
\end{table}

\subsection{Unseen Tasks}
\label{app sec:unseen}
We additionally evaluate generalization to unseen-task benchmarks that probe transfer beyond the training-aligned evaluation suite. AI2D~\cite{kembhavi2016diagram} measures diagram understanding and visual reasoning over structured scientific illustrations. ChartQA~\cite{masry2022chartqa} evaluates reasoning over charts and plots, requiring models to interpret visualized quantitative information. DocVQA~\cite{mathew2021docvqa} tests document understanding from visually rich document images, while InfoVQA~\cite{mathew2022infographicvqa} focuses on question answering over infographics that combine text, layout, and graphical elements. NaturalBench~\cite{li2024naturalbench} is designed to assess robust multimodal reasoning under more natural and diverse real-world inputs. RealWorldQA~\cite{xai2024grok15v} further examines practical visual question answering ability on real-world images. Finally, CMMMU~\cite{zhang2024cmmmu} extends evaluation to a broad, college-level multimodal benchmark that tests knowledge-intensive reasoning across diverse subjects. Together, these unseen tasks provide a complementary measure of out-of-distribution generalization and task transfer.

\begin{algorithm*}[t]
\caption{MAGIC Data Selection}
\label{alg:magic_select}
\small
\textbf{Input:} dataset $\mathcal{D}=\{x_i\}_{i=1}^N$, multimodal gain $\{g_i\}_{i=1}^N$, bridging relevance $\{b_i\}_{i=1}^N$, skill signatures $\{\phi_i\}_{i=1}^N$, target budget $M$, keep ratio $\rho$, shortlist factor $\eta$, weights $(\alpha,\beta)$, temperature $\tau$, bucket cap $\gamma$
\textbf{Output:} coreset $\mathcal{C}$
\begin{algorithmic}[1]
\STATE $\hat g_i \leftarrow \mathrm{Norm}(g_i),\quad \hat b_i \leftarrow \mathrm{Norm}(b_i),\quad i=1,\dots,N$ \hfill $\textcolor{cyan}{\triangleright}$ \textcolor{cyan}{\mbox{robust normalization}}
\STATE $q_i \leftarrow \alpha \hat g_i+\beta \hat b_i,\quad i=1,\dots,N$ \hfill $\textcolor{cyan}{\triangleright}$ \textcolor{cyan}{\mbox{joint quality score}}
\STATE $\mathcal{E}\leftarrow \mathrm{Top}_{\lceil \rho N\rceil}(\{g_i\}_{i=1}^N)$ \hfill $\textcolor{cyan}{\triangleright}$ \textcolor{cyan}{\mbox{multimodal-gain eligibility filter}}
\STATE $\mathcal{S}\leftarrow \mathrm{Top}_{\min(\lceil \eta M\rceil,|\mathcal{E}|)}(\{q_i\}_{i\in\mathcal{E}})$ \hfill $\textcolor{cyan}{\triangleright}$ \textcolor{cyan}{\mbox{quality shortlist}}
\STATE Partition $\mathcal{S}$ into buckets $\mathcal{B}=\{B_1,\dots,B_K\}$ by $\phi_i$ \hfill $\textcolor{cyan}{\triangleright}$ \textcolor{cyan}{\mbox{behavioral grouping}}
\STATE $m_j \leftarrow \sum_{i\in B_j}\exp(q_i/\tau),\quad j=1,\dots,K$ \hfill $\textcolor{cyan}{\triangleright}$ \textcolor{cyan}{\mbox{bucket mass}}
\STATE $p_j \leftarrow m_j / \sum_{r=1}^K m_r,\quad j=1,\dots,K$ \hfill $\textcolor{cyan}{\triangleright}$ \textcolor{cyan}{\mbox{normalized bucket weight}}
\STATE $n_j \leftarrow \min\!\left(|B_j|,\left\lceil \gamma M\right\rceil,\left\lfloor Mp_j\right\rfloor\right),\quad j=1,\dots,K$ \hfill $\textcolor{cyan}{\triangleright}$ \textcolor{cyan}{\mbox{capped initial quota}}
\STATE Redistribute $M-\sum_{j=1}^K n_j$ by descending fractional remainder of $Mp_j$ \hfill $\textcolor{cyan}{\triangleright}$ \textcolor{cyan}{\mbox{budget correction}}
\STATE $\mathcal{C}\leftarrow \bigcup_{j=1}^K \mathrm{Top}_{n_j}(\{q_i\}_{i\in B_j})$ \hfill $\textcolor{cyan}{\triangleright}$ \textcolor{cyan}{\mbox{intra-bucket top-score selection}}
\STATE Backfill from $\mathcal{S}\setminus\mathcal{C}$, then $\mathcal{E}\setminus\mathcal{C}$, by descending $q_i$ until $|\mathcal{C}|=M$ \hfill $\textcolor{cyan}{\triangleright}$ \textcolor{cyan}{\mbox{fill unused budget}}
\STATE \textbf{return }\(\mathcal{C}\)
\end{algorithmic}
\end{algorithm*}

\subsection{Coverage in the MAGIC Feature Space}

\begin{figure}[t]
\centering
    \includegraphics[width=1.0\linewidth]{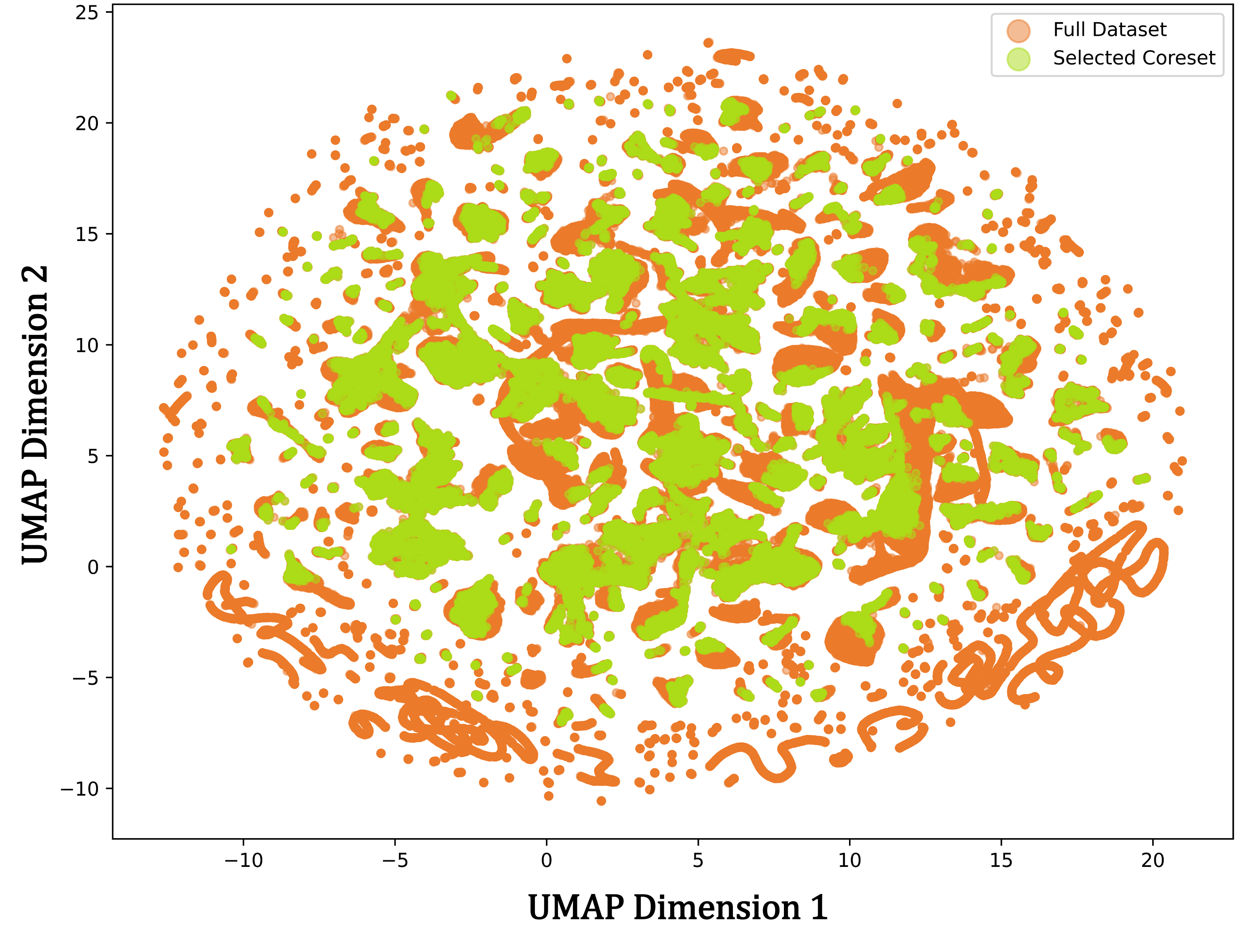}
    \vspace{-5pt}
    \caption{\textbf{UMAP visualization of the MAGIC feature space.} Each point represents one training sample embedded from its joint MAGIC descriptor, formed from Multimodal Gain (MG), Bridging Relevance (BR), and the processed Skill-Neuron (SN) signature. MAGIC preserves broad global coverage and local behavioral structure while selecting a compact subset.}
    \vspace{-6pt}
    \label{fig:coverage}
\end{figure}

To examine whether MAGIC preserves broad coverage of the training distribution, we visualize the joint MAGIC feature space using UMAP in Figure~\ref{fig:coverage}. Each point corresponds to one training sample embedded from its combined MAGIC descriptor, formed from Multimodal Gain (MG), Bridging Relevance (BR), and the processed Skill-Neuron (SN) signature. The orange points denote the full dataset, while the green points denote the selected coreset.

Two observations are notable. First, the selected coreset remains well distributed across the global support of the full dataset rather than collapsing onto a small number of dense regions. This suggests that MAGIC does not behave like a purely score-based filter that concentrates only on a narrow subset of samples. Second, the coreset also preserves many of the local structures present in the full dataset, indicating that the proposed skill-signature bucketing retains diverse latent behavioral modes while still favoring high-quality samples. Together, these results provide qualitative evidence that MAGIC preserves feature coverage in the multimodal training space while constructing a substantially smaller subset.

\begin{figure*}[t]
\centering
    \includegraphics[width=1.0\linewidth]{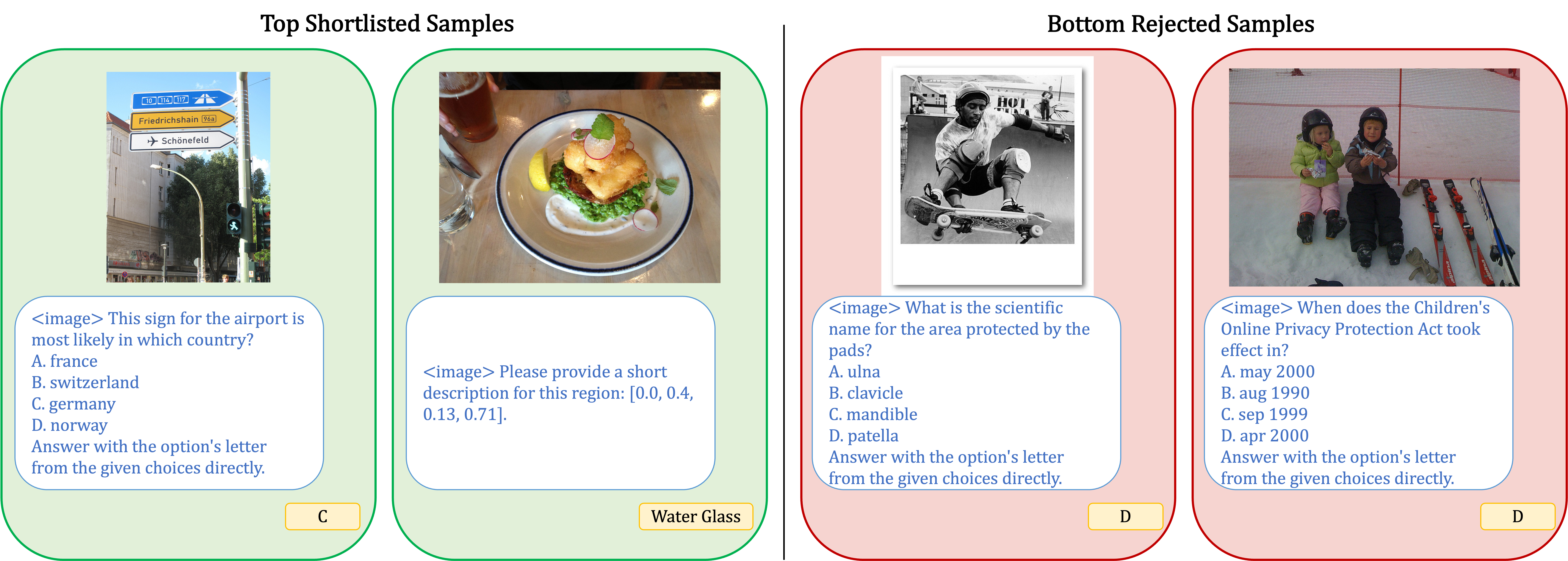} 
    \vspace{-8pt}
    \caption{\textbf{Qualitative examples of MAGIC selection.} The left two examples are highly ranked samples selected by MAGIC, while the right two are low-ranked samples rejected by the method. MAGIC prioritizes instances that require genuine visual grounding and multimodal reasoning, and deprioritizes samples that are largely answerable from language priors or provide weak visual supervision.}
    \vspace{-6pt}
    \label{fig:sample}
\end{figure*}

\subsection{Qualitative Examples of Selected and Rejected Samples}

Figure~\ref{fig:sample} provides qualitative examples of the samples prioritized and rejected by MAGIC. The \textbf{left} two examples are drawn from the top of the MAGIC ranking and illustrate the types of training instances favored by our selection criterion. These samples require genuine visual reasoning: the first example asks the model to infer the country from a road sign, which depends on recognizing localized textual and structural cues in the image, while the second requires grounded regional description from a specified image crop. Such examples are visually informative, strongly grounded, and behaviorally non-trivial, and therefore tend to receive high multimodal gain and bridging relevance scores.

In contrast, the \textbf{right} two examples are drawn from the bottom of the ranking and illustrate the types of samples that MAGIC deprioritizes. These examples are either largely recoverable from language priors or only weakly dependent on the image content. For instance, the question about the ``scientific name for the area protected by the pads'' can be answered primarily from textual world knowledge, while the question about the Children's Online Privacy Protection Act is unrelated to the visual evidence altogether. Such samples provide limited multimodal supervision and are therefore less useful for efficient visual instruction tuning.

Overall, these examples highlight that MAGIC favors samples where the image materially contributes to the answer and where the required computation is both grounded and behaviorally distinctive.

\end{document}